\documentclass[preprint,5p,times,twocolumn,authoryear]{elsarticle}
\usepackage{framed} 
\usepackage{multicol} 
\usepackage{nomencl} 
\makenomenclature
\renewcommand*\nompreamble{\begin{multicols}{2}}
\renewcommand*\nompostamble{\end{multicols}}
\usepackage{amssymb}
\usepackage{arydshln} 
\usepackage{amsthm}
\usepackage{graphicx}
\usepackage{subfig}
\usepackage{color}
\usepackage{url}
\usepackage[utf8]{inputenc}
\usepackage{natbib}
\definecolor{lightblue}{RGB}{0,176,240}
\definecolor{purple}{RGB}{153,102,255}
\usepackage{amsmath}
\usepackage[scr=boondox,  
            cal=esstix]   
           {mathalpha}
\usepackage{algorithm}
\usepackage{algpseudocode}
\usepackage{float}
\usepackage{multirow}
\usepackage[utf8]{inputenc}
\usepackage[colorlinks,linkcolor=blue, anchorcolor=blue, citecolor=blue]{hyperref}
\usepackage[switch]{lineno}
\journal{Elsevier}

\begin{document}

\begin{frontmatter}
\title{AeroReformer: Aerial Referring Transformer for UAV-based Referring Image Segmentation}

\author{Rui Li}
\ead{rui.li.8@warwick.ac.uk}
\author{Xiaowei Zhao\corref{cor2}}
\ead{xiaowei.zhao@warwick.ac.uk}
\address[label1]{Intelligent Control \& Smart Energy (ICSE) Research Group, School of Engineering, University of Warwick, Coventry, CV4 7AL, UK}
\cortext[cor2]{Corresponding author}

\begin{abstract}
As a novel and challenging task, referring segmentation combines computer vision and natural language processing to localise and segment objects based on textual descriptions. While Referring Image Segmentation (RIS) has been extensively studied in natural images, little attention has been given to aerial imagery, particularly from Unmanned Aerial Vehicles (UAVs). The unique challenges of UAV imagery, including complex spatial scales, occlusions, and varying object orientations, render existing RIS approaches ineffective. A key limitation has been the lack of UAV-specific datasets, as manually annotating pixel-level masks and generating textual descriptions is labour-intensive and time-consuming. To address this gap, we design an automatic labelling pipeline that leverages pre-existing UAV segmentation datasets and the Multimodal Large Language Models (MLLM) for generating textual descriptions. Furthermore, we propose Aerial Referring Transformer (AeroReformer), a novel framework for UAV Referring Image Segmentation (UAV-RIS), featuring a Vision-Language Cross-Attention  Module (VLCAM) for effective cross-modal understanding and a Rotation-Aware Multi-Scale Fusion (RAMSF) decoder to enhance segmentation accuracy in aerial scenes. Extensive experiments on two newly developed datasets demonstrate the superiority of AeroReformer over existing methods, establishing a new benchmark for UAV-RIS. The datasets and code will be publicly available at \url{https://github.com/lironui/AeroReformer}.
\end{abstract}

\end{frontmatter}


\section{Introduction}
\label{sec:1}

Referring Image Segmentation (RIS) aims to segment target objects in an image based on natural language expressions that describe their attributes or context \citep{li2018referring, ding2022vlt}. Unlike traditional image segmentation methods that rely on predefined semantic labels and operate within a constrained set of categories \citep{simonyan2014very, ronneberger2015u, wang2022unetformer}, referring image segmentation enables open-domain segmentation by utilising free-form textual descriptions as guidance \citep{hu2016segmentation, liu2017recurrent, lai2024lisa}. This capability significantly expands its applicability, allowing for more flexible and context-aware interpretation of imagery. In terms of the aerial scenario, UAV Referring Image Segmentation (UAV-RIS) has broad applications in domains such as text-guided environmental monitoring \citep{sharma2022uav}, land cover classification \citep{mienna2022land}, precision agriculture \citep{tahir2023application}, urban planning \citep{shao2021assessing} and risk assessment \citep{trepekli2022uav}, where identifying and segmenting specific objects or regions based on natural language descriptions is crucial. By leveraging the multimodal integration of vision and language, UAV-RIS can enhance the precision and adaptability of spatial analysis, making it easier to extract detailed, context-specific information from complex aerial imagery.

\begin{figure}[!h]
\centering
\includegraphics[width=9cm]{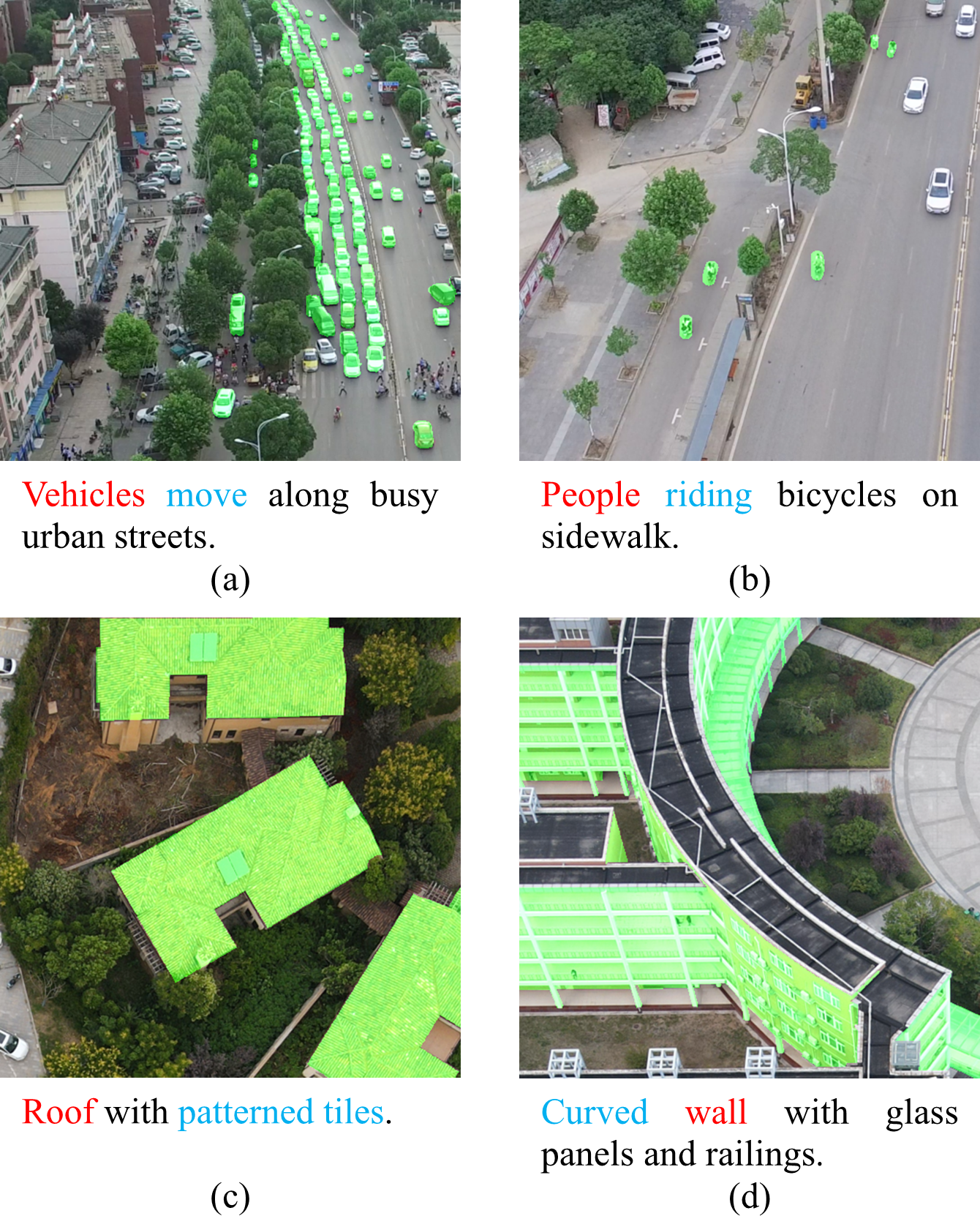}
\caption{The shifting viewpoints and varying
scene contexts are very common for UAV-captured images.}
\label{fig:1}
\end{figure}

Recently, benefit from the open-source datasets including RefSegRS \citep{yuan2024rrsis}, RRSIS-D \citep{liu2024rotated} and RISBench \citep{dong2024cross}, the Referring Remote Sensing Image Segmentation (RRSIS) has attracted more and more attentions \citep{lei2024exploring, shi2025multimodal, zhang2025referring, chen2025rsrefseg, li2025scale}. Despite these promising advances, UAV-RIS poses additional complexities due to the lower altitudes and agile motion of UAV platforms, leading to more pronounced occlusions, rapidly shifting viewpoints, and varying scene contexts \citep{lyu2020uavid, li2024lswinsr, zhang2023efficient}. Moreover, building a large-scale, high-quality UAV dataset for referring segmentation remains labour-intensive, underscoring the need for automated or semi-automated approaches to annotation. Therefore, it remains under-investigated in aerial imagery, especially those captured by UAVs. 

This paper addresses the challenges of UAV-RIS by introducing UAV-specific datasets and framework, expanding the scope of referring image segmentation to UAV imagery and establishing a foundation for future research. Specifically, we develop an automatic labelling framework that leverages open-source and pre-existing UAV segmentation datasets along with a Multimodal Large Language Model (MLLM). In our data generation process, segmentation masks are obtained from existing dataset annotations, and a cropped image paired with a well-designed prompt is fed into the MLLM to generate textual descriptions of the target object. This approach streamlines the annotation and description generation process, reducing the time and effort required for manual labelling.   

Meanwhile, UAV imagery presents unique challenges compared to natural images, including significant scale variations, diverse object orientations, and complex background clutter \citep{lyu2020uavid, li2024lswinsr, zhang2023efficient}. To effectively bridge the gap between visual and linguistic modalities, we propose a UAV-specific RIS model, \textit{i.e.} AeroReformer, featuring a Vision-Language Cross-Attention Module (VLCAM) for robust cross-modal understanding and a Rotation-Aware Multi-Scale Fusion (RAMSF) decoder to address spatial variations in UAV imagery. VLCAM dynamically aligns visual features with linguistic queries, ensuring that textual descriptions are accurately mapped to corresponding image regions, even under complex conditions such as occlusions and scale variations. Meanwhile, RAMSF enhances the segmentation process by incorporating rotation-aware convolutions and multi-scale feature aggregation, preserving orientation consistency while maintaining high-resolution spatial details. The integration of these two modules enables our model to outperform existing methods, achieving state-of-the-art results on UAVid-RIS and VDD-RIS, two datasets generated by our proposed pipeline. The main contribution of this paper can be summarised as:

\begin{enumerate}[(1)]
\item An automatic dataset generation framework is designed, enabling the transformation of labelled segmentation datasets into their LLM-aided counterparts.

\item Two UAV-RIS datasets, UAVid-RIS and VDD-RIS, are constructed from open-source datasets, providing a benchmark for UAV-RIS research and evaluation.

\item A novel UAV-RIS network, AeroReformer, is designed, incorporating a Vision-Language Cross-Attention Module (VLCAM) and a Rotation-Aware Multi-Scale Fusion (RAMSF) decoder, achieving state-of-the-art performance on UAVid-RIS and VDD-RIS.

\end{enumerate}

The remainder of this paper is organized as follows: Section \ref{sec:2} reviews related work. Section \ref{sec:3} presents the UAV referring segmentation dataset generation pipeline. Section \ref{sec:4} introduces the proposed AeroReformer model, explaining its architecture. Section \ref{sec:5} describes the experimental setup, dataset details, and evaluation metrics and presents a performance analysis. Finally, Section \ref{sec:6} concludes the research and discusses potential directions for future research.

\section{Related Work}
\label{sec:2}

\subsection{Referring Image Segmentation for Natural Images}
\label{sec:2.1}
RIS is a fundamental task in vision-language understanding, where the goal is to segment objects in an image based on a given natural language expression \citep{li2018referring, yu2016modeling, yu2018mattnet}. This task demands a fine-grained alignment between textual descriptions and visual features to correctly localize and delineate the referenced objects. Compared to conventional segmentation tasks that rely on predefined categories, RIS enables a more flexible and user-specific segmentation process.

In the early stages, initial RIS models relied primarily on Convolutional Neural Networks (CNNs) to extract visual features and Recurrent Neural Networks (RNNs) to process textual descriptions \citep{li2018referring, hu2016segmentation, nagaraja2016modeling}. These models performed feature fusion by concatenating visual and linguistic representations before feeding them into a segmentation head. Specifically, \cite{hu2016segmentation} first introduced RIS to address the limitations of traditional semantic segmentation when handling complex textual descriptions. Later, \cite{li2018referring} and  \cite{nagaraja2016modeling} explored bidirectional interactions between visual and textual features, improving the multimodal understanding of objects through structured representations. Further advancements introduced dynamic multimodal networks, such as the work by \cite{margffoy2018dynamic}, which incorporated recursive reasoning mechanisms to enhance the integration of linguistic and visual information.

As RIS models evolved, researchers recognised the importance of cross-modal feature alignment, leading to the introduction of attention-based strategies \citep{shi2018key, ye2019cross, hu2020bi}. For example, \cite{shi2018key} introduced a keyword-aware segmentation model, refining object-region relationships based on key linguistic cues. These approaches significantly improved object localisation and contextual interpretation in RIS tasks.  \cite{ye2019cross} proposed a cross-modal self-attention module to capture long-range dependencies between textual and visual elements, improving multimodal fusion. Similarly, \cite{hu2020bi} developed a bidirectional cross-modal attention mechanism, enabling deeper interaction between the modalities.

The recent emergence of Transformer-based architectures has significantly advanced RIS, offering global modelling capabilities and superior multimodal integration. Unlike CNN-based methods, which rely on local receptive fields, Transformers enable long-range dependencies and self-attention mechanisms, making them particularly effective for RIS \citep{ding2022vlt, yang2022lavt, liu2023gres}. Several notable works have leveraged this architecture. VLT designed a query-based Transformer framework, enriching textual comprehension by dynamically generating language query embeddings \citep{ding2022vlt}. LAVT proposed language-aware attention mechanisms to enhance early fusion between the two modalities, enabling more precise segmentation \citep{yang2022lavt}. GRES further refined multimodal alignment by explicitly modelling dependencies between different textual tokens and visual regions, leading to more robust segmentation performance \citep{liu2023gres}.

\subsection{ Referring Remote Sensing Image Segmentation}
\label{sec:2.2}
Referring Remote Sensing Image Segmentation (RRSIS) is a specialized task that aims to extract pixel-wise segmentation masks from remote sensing imagery based on natural language expressions \citep{yuan2024rrsis, liu2024rotated}. While it has significant applications in environmental monitoring, land cover classification, disaster response, and urban planning \citep{sun2022visual, li2024language}, progress in this field hinges critically on suitable datasets that capture the complexity of remote sensing imagery. One of the earliest datasets was RefSegRS, introduced by LGCE \citep{yuan2024rrsis}, which enabled initial efforts to adapt RIS methods from natural images to the remote sensing domain. To enhance the diversity and improve the generalizability of trained models, \cite{liu2024rotated} proposed RRSIS-D, a substantially larger dataset for benchmarking mainstream RIS models in remote sensing image segmentation. More recently, RISBench, has also been introduced to further advance the development and evaluation of RRSIS methods. 

Building on these datasets, recent RRSIS research has explored strategies to address scale variations, complex backgrounds, and orientation diversity. LGCE \citep{yuan2024rrsis} pioneered a language-guided cross-scale enhancement module to fuse shallow and deep features for improved segmentation accuracy, whereas \cite{liu2024rotated} proposed the Rotated Multi-Scale Interaction Network (RMSIN), which integrates intra-scale and cross-scale interactions alongside rotated convolutions to better handle directional variations. Beyond scale-aware models, \cite{pan2024rethinking} analysed the implicit optimization mechanisms in existing models and proposed an explicit affinity alignment approach, incorporating a new loss function to improve textual-visual feature interaction. More recent studies have introduced refined image-text alignment strategies to improve RRSIS performance. To be specific, FIANet \citep{lei2024exploring} introduced a fine-grained alignment module with object-positional enhancement, integrating a text-aware self-attention mechanism to refine segmentation accuracy. Similarly, CroBIM \citep{dong2024cross} leveraged a context-aware prompt modulation module, which optimizes post-fusion feature interactions and employs a mutual-interaction decoder to refine segmentation masks. Recently, SBANet \citep{li2025scale} introduced a bidirectional alignment mechanism and a scale-wise attention module to enhance mutual guidance between vision and language features, effectively refining segmentation masks in referring remote sensing image segmentation. BTDNet \citep{zhang2025referring} employs a bidirectional spatial correlation module and a target-background twin-stream decoder to improve multimodal alignment and fine-grained object differentiation, achieving improved segmentation performance.

\subsection{Visual Grounding for Aerial Images}
\label{sec:2.3}
Another active vision-and-language research in
remote sensing community is visual grounding for aerial images, focusing on localizing target objects within aerial scenes using natural language queries \citep{sun2022visual, zhao2021high, zhan2023rsvg}. In contrast to RRSIS, which demands detailed pixel-level masks, visual grounding is primarily concerned with identifying object-level regions—typically represented as bounding boxes \citep{sun2022visual}. This task leverages the unique characteristics of aerial imagery, where targets often exhibit complex spatial relationships and may not be visually prominent due to scale variations and cluttered backgrounds.

Early frameworks, such as GeoVG \citep{sun2022visual}, pioneered this approach by integrating a language encoder that captures geospatial relationships with an image encoder that adaptively attends to aerial scenes. By fusing these modalities, GeoVG established a one-stage process that effectively translates natural language cues into object localization. Building on this foundation, subsequent models have introduced advanced fusion strategies. For instance, modules like the Transformer-based Multi-Granularity Visual Language Fusion (MGVLF) \citep{zhan2023rsvg} exploit both multi-scale visual features and multi-granularity textual embeddings, resulting in more discriminative representations that address the challenges posed by large-scale variations and busy backgrounds. Vision-Semantic Multimodal Representation (VSMR) enhanced multi-level feature integration, refining how visual and textual features are jointly processed to improve localization robustness \citep{ding2024visual}. Further improvements have been achieved through progressive attention mechanisms. The Language-guided Progressive Visual Attention (LPVA) framework, for example, dynamically adjusts visual features at various scales and levels, ensuring that the visual backbone concentrates on expression-relevant information \citep{li2024language}. This is complemented by multi-level feature enhancement decoders, which aggregate contextual information to boost feature distinctiveness and suppress irrelevant regions.

\section{UAV Referring Segmentation Dataset Generation}
\label{sec:3} 

Although several RRSIS datasets have been introduced, they predominantly focus on vertically captured (nadir-view) satellites and aerial imagery, limiting their applicability to UAV-based scenarios. Unlike satellite imagery, UAVs operate at lower altitudes with dynamic viewing angles, resulting in significant variations in object appearance due to oblique perspectives, occlusions, and scale distortions. These factors make existing RRSIS datasets insufficient for UAV-RIS tasks, where diverse viewpoints and fine-grained scene details are crucial for accurate segmentation. To address this gap, we introduce a UAV-RIS dataset generation pipeline, ensuring more realistic and comprehensive benchmarking for UAV-based vision-language tasks.

\begin{figure*}[]
\centering
\includegraphics[width=18cm]{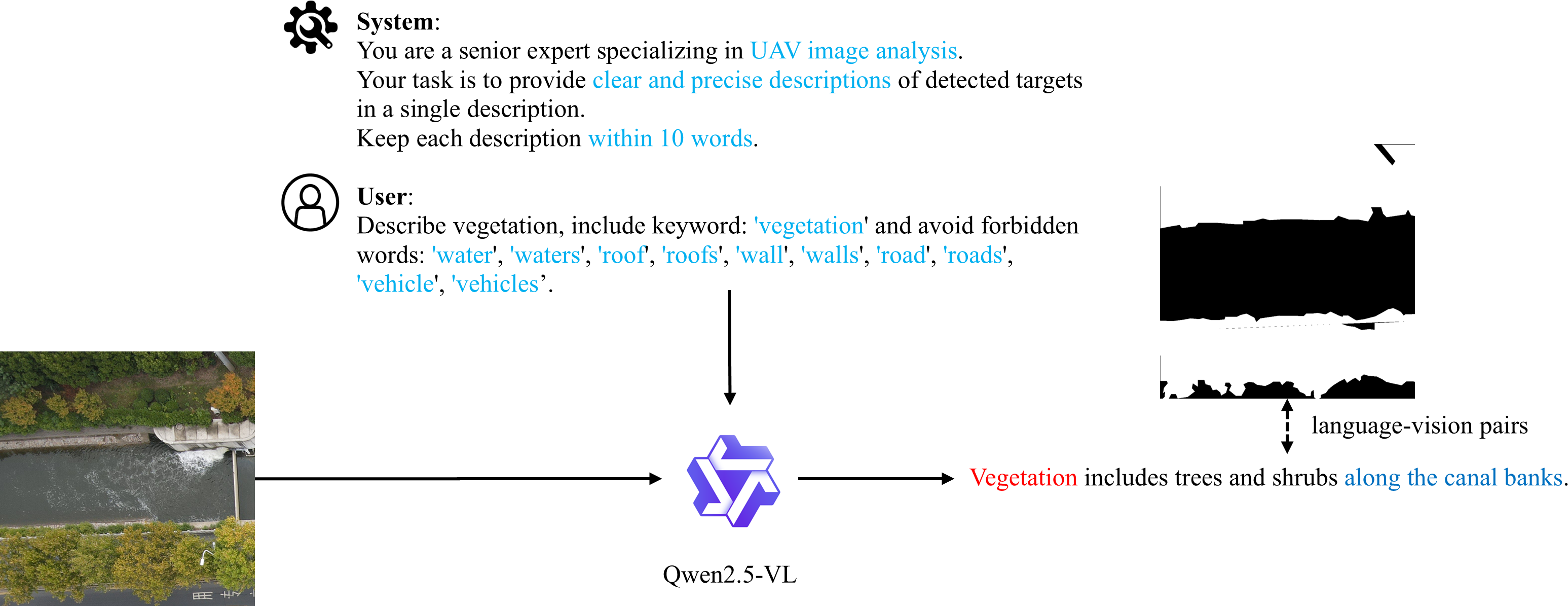}
\caption{The automatic UAV referring segmentation dataset generation pipeline.}
\label{fig:2}
\end{figure*}

\subsection{Dataset Construction and Analysis}

To advance UAV-RIS, we present a fully automated pipeline that leverages pre-existing UAV segmentation datasets. Unlike traditional approaches requiring manual annotations, our pipeline efficiently generates language-vision pairs by integrating segmentation masks with large language models. The data generation process, as shown in Fig. \ref{fig:2}, is structured as follows:

\begin{figure}[!h]
\centering
\includegraphics[width=9cm]{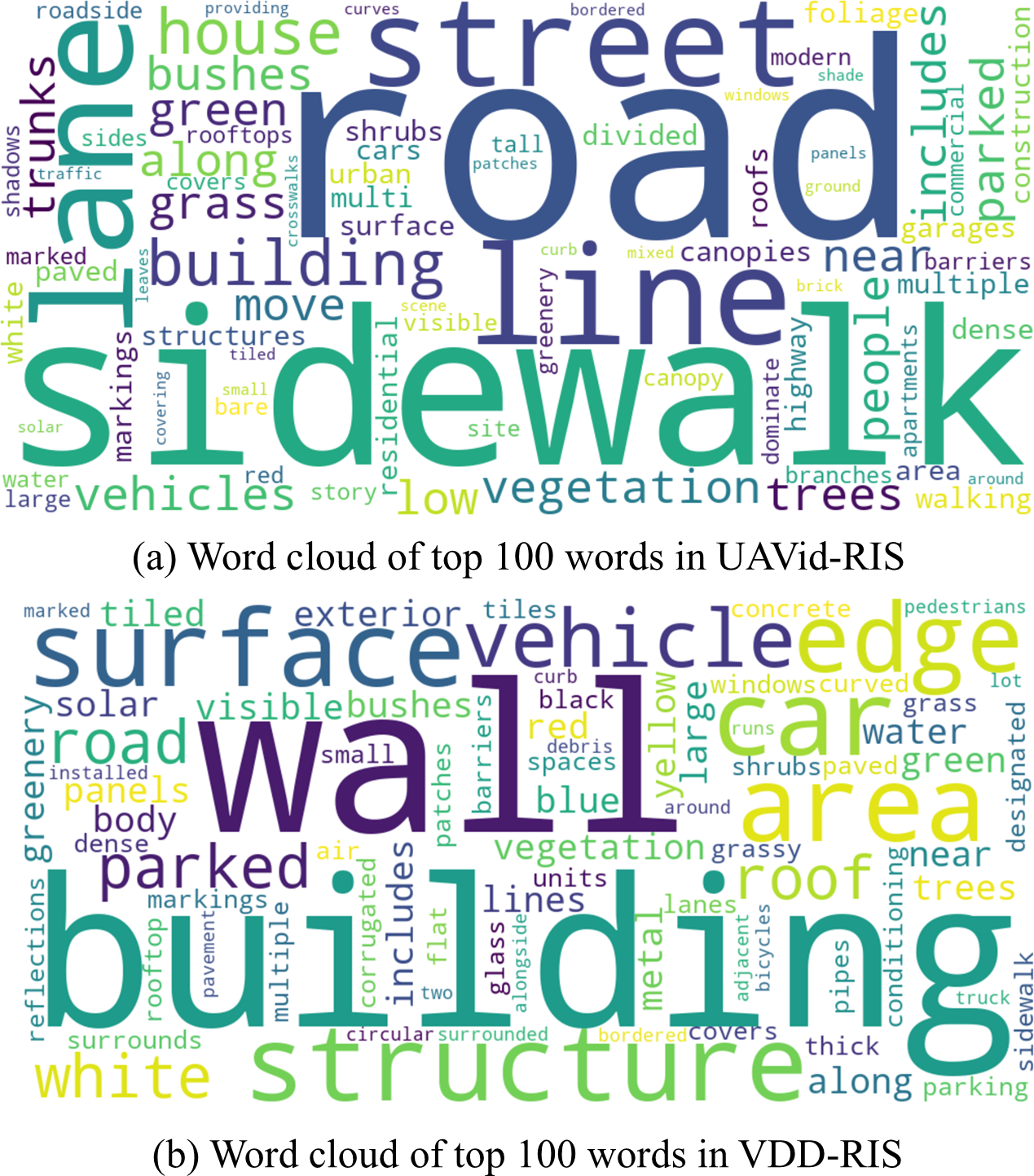}
\caption{Word cloud for top 100 words within the expressions of (a) UAVid-RIS and (b) VDD-RIS.}
\label{fig:3}
\end{figure}

\begin{figure*}[!h]
\centering
\includegraphics[width=18cm]{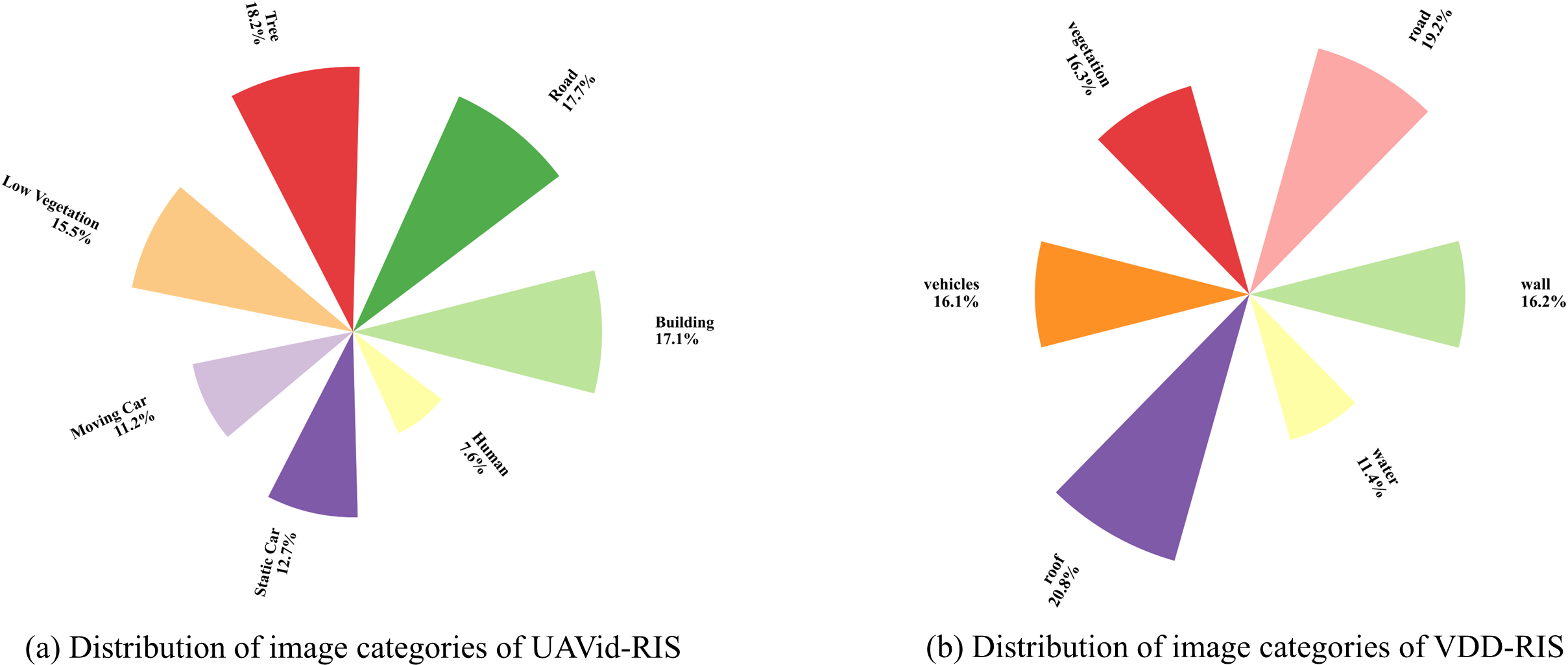}
\caption{Distribution of image categories of (a) UAVid-RIS and (b) VDD-RIS.}
\label{fig:4}
\end{figure*}

\begin{itemize}
    \item \textbf{Step 1: Segmentation-based Cropping.}  
    
    Given a pre-labelled UAV segmentation dataset, we first partition images and masks according to their object categories to $ 1024 \times 1024 $ patches. To ensure meaningful segmentation patches, we apply a filtering mechanism based on class presence and distribution. For each patch, the class distribution is computed by analysing the pixel proportions of predefined categories such as buildings, trees, roads, and vehicles. Patches containing only a single dominant class (occupying more than 70\% of the patch) or those with minimal class representation (below predefined thresholds) are discarded. This step enhances dataset diversity and ensures the presence of multiple meaningful classes in each selected patch.

    \item \textbf{Step 2: Vision-Language Description Generation.}  
    
    Once cropped, the segmented regions are processed using Qwen2.5-VL 7B \citep{bai2025qwen2}, a vision-language model capable of generating concise and context-aware descriptions for detected objects. The model takes as input the image patch along with a predefined prompt specifying the target object class. Each prompt ensures the inclusion of the relevant object category while avoiding semantically conflicting terms. The object categories vary depending on the pre-labelled segmentation dataset used. Each dataset's predefined classes are utilised to generate corresponding descriptions, ensuring alignment with its original annotations. Tailored instructions are provided for each category to produce accurate and concise textual descriptions. To maintain consistency, the model operates under a controlled generation setting where responses are constrained to a maximum of 10 words. Additionally, the prompt explicitly instructs the model to avoid irrelevant or misleading terms, ensuring that the generated descriptions remain semantically aligned with the visual content. The processed text-image pairs form the basis of the final dataset, enabling high-quality referring segmentation in UAV imagery.

    \item \textbf{Step 3: Automatic Description Refinement.}
    
    To maintain dataset consistency and clarity, we implement a text-cleaning process that removes ambiguous or uninformative phrases, such as \textit{``no visible"}. This automatic post-processing step ensures that all descriptions remain meaningful and directly correspond to the visual content of the segmented region. Finally, the annotations are formatted to align with the RefCOCO \citep{lin2014microsoft} dataset structure, enhancing compatibility with existing referring segmentation models and facilitating seamless integration into vision-language benchmarks.
    
\end{itemize}

Our proposed pipeline has been applied to two widely used UAV segmentation datasets, UAVid \citep{lyu2020uavid} and VDD \citep{cai2023vdd}, to generate their corresponding referring segmentation versions, namely UAVid-RIS and VDD-RIS. By leveraging the pre-existing segmentation annotations, our approach automatically extracts meaningful patches and generates language descriptions for each segmented region. This transformation enables the datasets to be directly used for vision-language tasks, expanding their applicability. 

To gain insights into the linguistic and semantic characteristics of the generated dataset, we present a word cloud of the 100 most frequent words in Fig. \ref{fig:3}, offering an overview of the variety of object descriptions. Additionally, Fig. \ref{fig:4} shows the image category distribution, providing a general understanding of the dataset composition and the occurrence of different object categories in UAVid-RIS and VDD-RIS.

\subsection{Advantages and Disadvantages of the Designed Pipeline}

The designed pipeline offers several advantages that make it highly effective for UAV-RIS tasks, including:

\begin{itemize}
    \item \textbf{Fully Automated Process.} 
    
    The dataset generation pipeline is entirely automatic, eliminating the need for manual annotations. This significantly reduces human effort and makes it highly scalable for large-scale datasets.
    
    \item \textbf{Leverages Pre-existing Datasets.} 
    
    Instead of requiring new annotations, the method takes advantage of already labelled segmentation datasets, making it efficient and cost-effective.
    
    \item \textbf{Multi-Label Representation.} 
    
    A single image can have multiple referring expressions since it may contain multiple object categories, providing a richer semantic understanding and enabling a more comprehensive interpretation of the scene.

    \item \textbf{Scalable for Large Datasets.}
    
    The fully automatic pipeline allows for the generation of large-scale datasets without significant computational overhead, making it ideal for deep learning applications that require vast amounts of training data.

     \item \textbf{Diverse Language Descriptions.}
     
    Since the dataset’s descriptions are generated by a large language model, it provides various expressions for the same object category. This enhances the dataset’s linguistic diversity, making it more robust for vision-language models.
    
\end{itemize}

\begin{figure*}[!h]
\centering
\includegraphics[width=18cm]{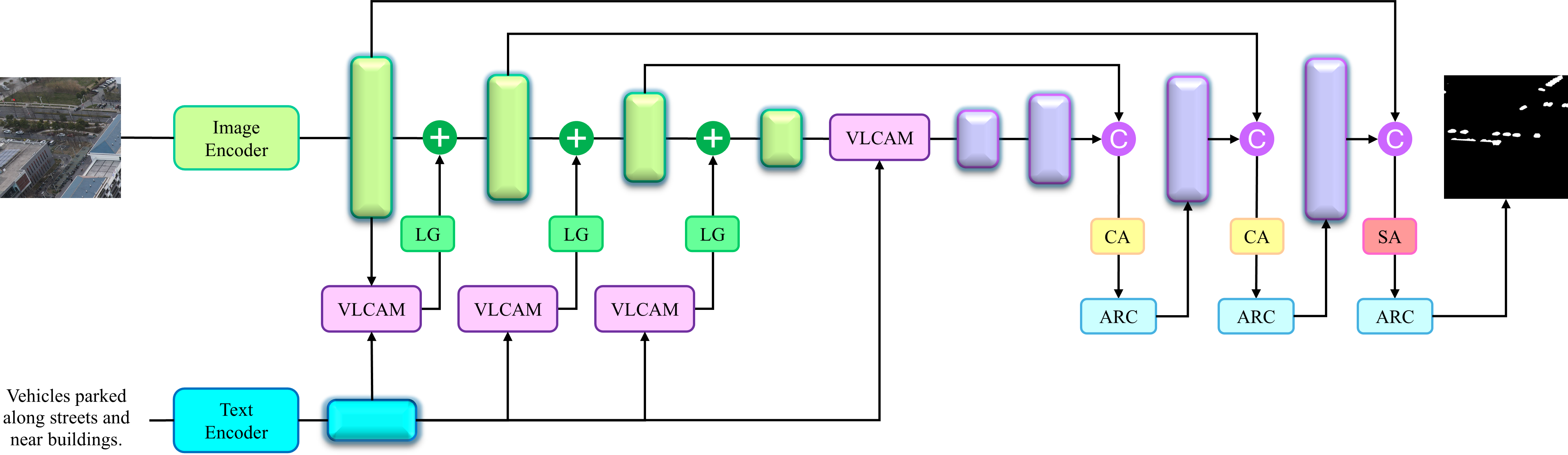}
\caption{Overview of the proposed AeroReformer.}
\label{fig:5}
\end{figure*}

At the same time, there are also certain limitations that should be considered including:

\begin{itemize}
    \item \textbf{Lack of Fine-Grained Object Features.}  
    
    The descriptions generated by the language model focus on object categories rather than detailed attributes such as colour, texture, shape, or exact dimensions. This may limit the dataset's applicability for fine-grained vision tasks. Additionally, although semantically conflicting words are explicitly excluded in the prompt, the model may still occasionally include them in the generated text, leading to potential inconsistencies in the descriptions.
    
    \item \textbf{Limited Spatial and Positional Context.}  
    
    While the dataset retains spatial and positional information, it is provided at the class level rather than for individual objects. Since the annotations correspond to entire object categories rather than distinct instances, precise localisation of single objects is not explicitly available.
    
    \item \textbf{Data Quality Depends on the Pre-Existing Dataset.}  
    
    The overall quality of the generated dataset is inherently dependent on the accuracy and granularity of the segmentation dataset. If the segmentation labels are noisy, incomplete, or overly coarse, it may negatively impact the quality of the generated language descriptions and corresponding annotations.

\end{itemize}

Despite these limitations, the generated dataset provides a scalable and efficient solution for vision-language learning in UAV imagery, making it a valuable resource for automatic referring segmentation. As the first publicly available dataset for UAV-RIS, it establishes a foundational benchmark for future research in this field. All data generation code will be openly released to facilitate research and drive advancements in the remote sensing community.

\section{Methodology}
\label{sec:4}

\subsection{Problem formulation}
\label{sec:4.1}

This study aims to tackle the challenge of referring image segmentation in UAV imagery, where the objective is to generate an accurate segmentation mask for a target category based on a given natural language description. Formally, let \( I \in \mathbb{R}^{H \times W \times C} \) denote an aerial image, where \( H \), \( W \), and \( C \) correspond to the image height, width, and number of channels, respectively. A textual query \( T = \{t_1, t_2, \dots, t_N\} \) serves as the semantic reference, where \( N \) represents the number of words or tokens in the description.  

The goal is to predict a binary segmentation mask \( O \in \{0,1\}^{H \times W} \), where each pixel \( p \in I \) is classified as either belonging to the category described by \( T \) or not. Given a dataset \( \Omega = \{(I_i, T_i, G_i)\}_{i=1}^{Num} \), where \( G_i \in \{0,1\}^{H \times W} \) represents the corresponding ground truth mask and \( Num \) is the total number of samples, the objective is to develop a function \( f \) that maps the image-text pair \( (I, T) \) to \( O \) by effectively learning cross-modal associations between linguistic descriptions and visual features.  

\subsection{Overview of the architecture}
\label{sec:4.2}

The overall architecture of our proposed AeroReformer is depicted in Fig.~\ref{fig:5}.  Our AeroReformer builds upon LAVT \citep{yang2022lavt}, maintaining its encoders \emph{e.g.,}~Swin Transformer \citep{liu2021swin} for extracting multi-modal inputs while improving vision-language fusion and mask prediction. Meanwhile, we propose a vision-language cross-attention fusion module, enhancing the interaction between visual and linguistic features. Additionally, we introduce a rotation-aware multi-scale fusion decoder, allowing better adaptation to aerial imagery with varying orientations.

\begin{figure*}[]
\centering
\includegraphics[width=15cm]{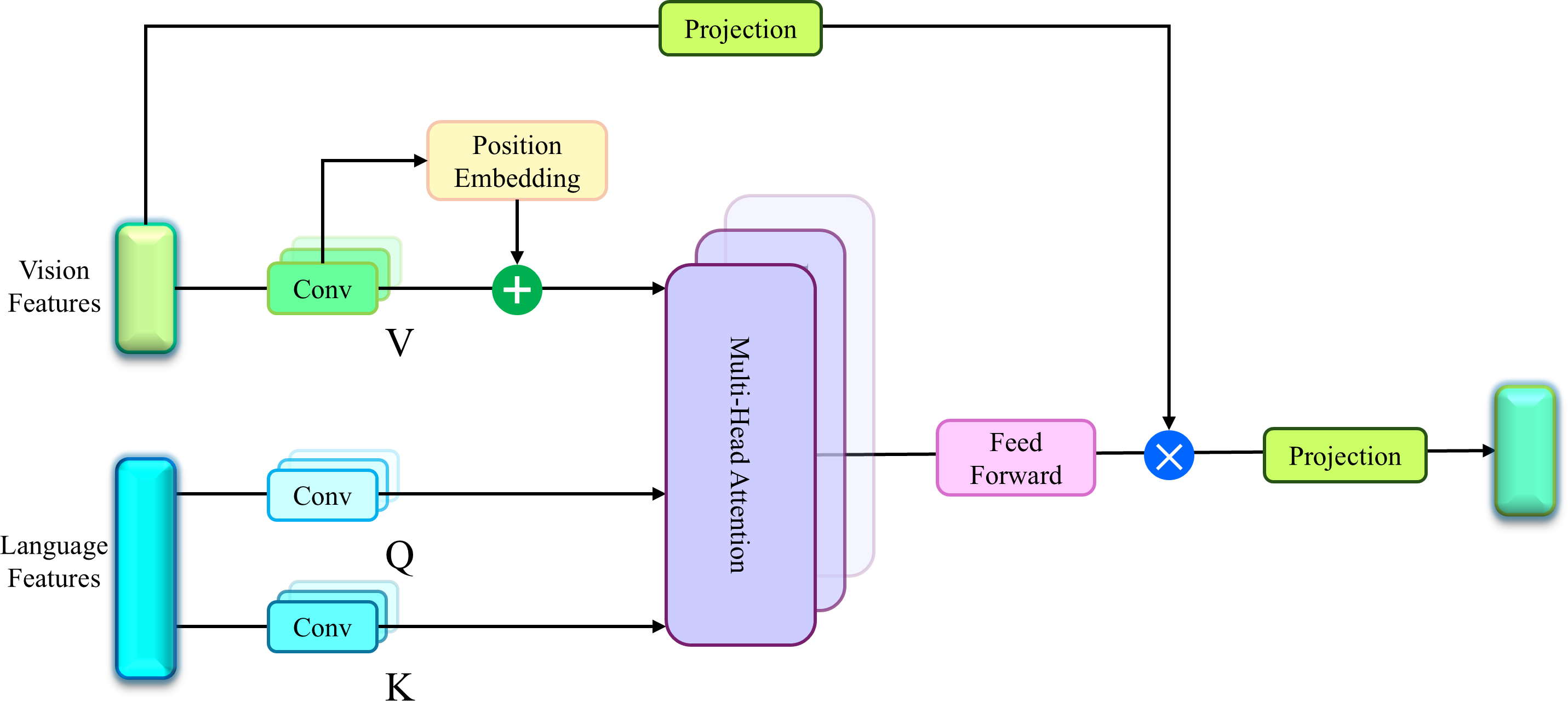}
\caption{Pipeline of the Vision-Language Cross-Attention Module (VLCAM).}
\label{fig:6}
\end{figure*}

\subsubsection{Vision-language cross-attention fusion module}

To effectively integrate linguistic and visual information, we introduce a vision-language cross-attention fusion module that enhances cross-modal feature interaction, as shown in Fig. \ref{fig:6}. This module replaces the Pixel-Word Attention Module (PWAM) with a more structured mechanism that utilises multi-head cross-attention to improve feature alignment between textual and visual representations.

Given an input aerial image \( I \in \mathbb{R}^{H \times W \times C} \) and a natural language expression \( T = \{t_1, t_2, ..., t_N\} \), our model extracts corresponding visual and linguistic features using hierarchical encoders. Let \( F_V \in \mathbb{R}^{H'W' \times C_V} \) denote the visual features extracted from the image, where \( H'W' \) represents the spatial dimension of the feature map and \( C_V \) is the number of visual feature channels. Similarly, let \( F_L \in \mathbb{R}^{N \times C_L} \) represent the linguistic features extracted from the text, where \( N \) is the number of tokens, and \( C_L \) is the language feature dimension.

To facilitate cross-modal interactions between vision and language, we first project the features into a common embedding space using convolutional layers:

\begin{equation}
\begin{aligned}
    Q_V &= \mathscr{I} \big( \mathscr{C}_{1D} (F_V, C_K) \big), \\
    K_L &= \mathscr{C}_{1D} (F_L, C_K), \\
    V_L &= \mathscr{C}_{1D} (F_L, C_V).
\end{aligned}
\end{equation}
where \( \mathscr{C}_{1D} \) represents the 1D convolution operation, \( \mathscr{I} \) denotes instance normalization. \( C_K \) represents the key-query dimension, while \( C_V \) denotes the value dimension. The query features \( Q_V \) are extracted from vision features \( F_V \) using a 1D convolution followed by instance normalization. The key \( K_L \) and value \( V_L \) are computed from the language features \( F_L \) using separate 1D convolutional layers. The positional encoding is enabled for query features:

\begin{equation}
Q_V = Q_V + P_V,
\end{equation}
where \( P_V \) is a learnable positional encoding that provides spatial awareness to the vision features.

Next, we compute multi-head attention scores using scaled dot-product attention while each head can be defined as:

\begin{equation}
\text{Attn} = \text{softmax} \left( \frac{Q_V K_L^T}{\sqrt{C_K}} \right),
\end{equation}
where the dot product of queries and keys is scaled by \( \sqrt{C_K} \) to stabilise gradients and prevent extreme values. The attended visual-language features are then computed as:

\begin{equation}
F_{VL} = \text{Attn} \cdot V_L.
\end{equation}

The cross-modal representation \( F_{VL} \in \mathbb{R}^{H'W' \times C_V} \) is then fused with the original vision features via element-wise interaction:

\begin{equation} F_{Fused} = \mathscr{I} \big( \mathscr{C}_{1D} (F_{VL} \odot F_V, C_V) \big), \end{equation}
where \( \odot \) represents element-wise multiplication, generating enriched visual features that incorporate linguistic information. Instance normalisation is applied to stabilise feature distributions. Thereafter, a Feed-Forward Network (FFN) is applied, consisting of two convolutional layers with ReLU activation: 

\begin{equation}
F_{FFN} = \mathscr{I} \Big( \mathscr{C}_{1D} \big( \mathscr{D} ( \mathscr{R} (\mathscr{C}_{1D}(F_{Fused}, 4C_V)) ) , C_V \big) \Big).
\end{equation}

Finally, the fused representation undergoes a final projection and dropout operation:

\begin{equation}
F_{out} = \mathscr{D} (\mathscr{C}_{1D} (F_{FFN}, C_V)).
\end{equation}
where \( \mathscr{D} \) represents the dropout operation and \( \mathscr{R} \) denotes the ReLU activation function

\subsubsection{Rotation-aware Multi-Scale Fusion Decoder}

To take full advantage of extracted futures, the rotation-aware multi-scale fusion decoder is designed with multi-scale feature aggregation operations. Specifically, given a set of encoder feature maps at different scales:
\begin{equation}
\begin{aligned}
    X_1 &\in \mathbb{R}^{B \times C_1 \times \frac{H}{4}  \times \frac{W}{4}},  \\
    X_2 &\in \mathbb{R}^{B \times C_2 \times \frac{H}{8}  \times \frac{W}{8}},  \\
    X_3 &\in \mathbb{R}^{B \times C_3 \times \frac{H}{16} \times \frac{W}{16}}, \\
    X_4 &\in \mathbb{R}^{B \times C_4 \times \frac{H}{32} \times \frac{W}{32}}.
\end{aligned}
\end{equation}
where \( X_i \) represents the feature maps at different resolutions, and \( C_i \) denotes the corresponding number of channels. To maintain scale consistency, lateral transformations are applied using \( 1 \times 1 \) convolutions first:
\begin{equation}
L_i = \mathscr{R} (\mathscr{C}_{2D}(X_i)), \quad i \in \{1,2,3,4\}
\end{equation}
where \( \mathscr{C}_{2D} \) is 2D convolutions.

\begin{figure}[htb]
\centering
\includegraphics[width=8cm]{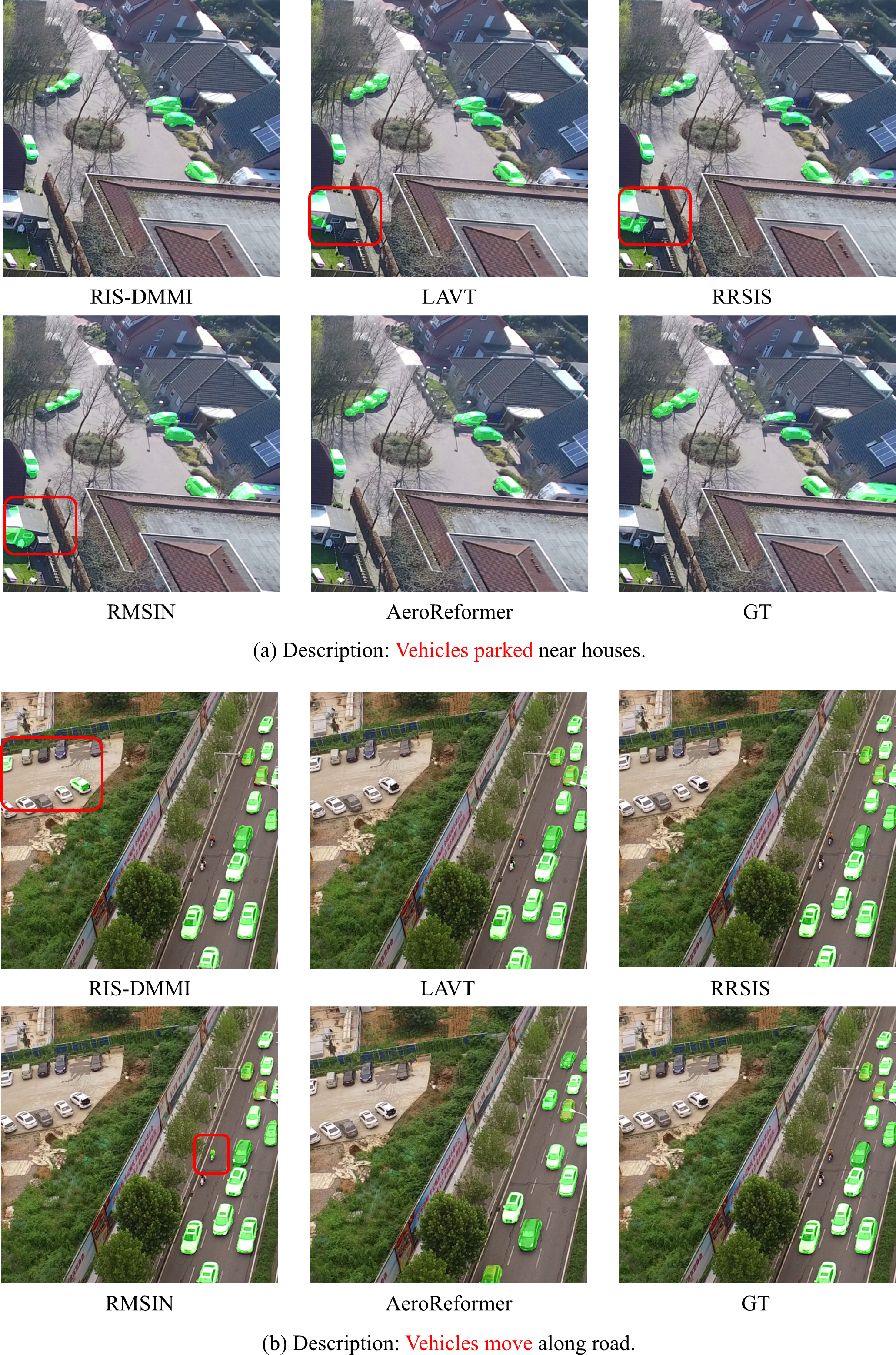}
\caption{Visual comparison of referring segmentation results on UAVid-RIS dataset.}
\label{fig:7}
\end{figure}

Thereafter, upsampled decoder feature maps \( Y_i \) are concatenated with the corresponding lateral feature maps obtained from the encoder:
\begin{equation}
F_i = \text{Concat}(L_i, Y_i), \quad i \in \{1,2,3\}
\end{equation}

This concatenated feature maps are then refined using either a channel attention module (\( i=2, 3 \)) or a spatial attention module (\( i=1 \)), which leverages L2 normalisation and adaptive weighting instead of traditional softmax-based dot-product attention, reducing both time and memory costs based on our previous work \citep{li2021multistage, li2021abcnet}. The refined feature representation is given by:

\begin{equation}
F_{\text{CA}}' = X + \gamma_c \cdot \mathrm{reshape} \left( \frac{\sum\limits_{n} V_{n} + Q \otimes (K \otimes V)}{H \times W + Q \otimes \sum\limits_{n} K_{n} + \varepsilon} \right),
\end{equation}

\begin{equation}
F_{\text{SA}}' = X + \gamma_s \cdot \mathrm{reshape} \left( \frac{\sum\limits_{c} V_{c} + Q \otimes (K \otimes V)}{H \times W + Q \otimes \sum\limits_{c} K_{c} + \varepsilon} \right).
\end{equation}
where \( X \) is the input feature map, \( Q, K, V \) are the query, key, and value matrices, respectively, obtained from learned projections of \( X \), and \( \otimes \) represents matrix multiplication. The terms \( \sum\limits_{n} V_n \) and \( \sum\limits_{c} V_c \) aggregate values across spatial and channel dimensions, respectively. \( \varepsilon \) ensures numerical stability. The scaling factors \( \gamma_c \) and \( \gamma_s \) are learnable parameters for channel and spatial attention mechanisms.

This formulation integrates both self-attention mechanisms, effectively capturing feature correlations across different dimensions while maintaining computational efficiency through L2-normalised weighting. Specifically, the channel attention module is applied to the first two fused feature maps, as they contain multiple channels representing hierarchical multi-scale information. This mechanism models dependencies between different channels, allowing the network to dynamically recalibrate inter-channel relationships and enhance contextual coherence. Meanwhile, for the final fused feature map, which is at the highest resolution and contains detailed spatial semantics, the spatial attention mechanism is employed. This ensures that long-range pixel dependencies are effectively captured, refining feature distributions across spatial dimensions. 

To further improve feature representation across different orientations, which frequently occur in UAV imagery, we incorporate the Adaptive Rotated Convolution (ARC) module \citep{pu2023adaptive} into the fused features. Unlike standard convolution, where a fixed kernel is applied to all inputs, ARC adapts its filters to align with the directional variations present in imagery. Specifically, given an input feature map \( X \), the routing function \( \mathscr{F} \) predicts a set of rotation angles \( \theta \) and corresponding weights \( \lambda \):

\begin{equation}
 \theta, \lambda = \mathscr{F}({X}).
\end{equation}
Each of the \( n \) convolution kernels \( W_i \) is then rotated according to its corresponding predicted angle:

\begin{equation}
  W^{\prime}_i = \text{Rotate}(W_i, \theta_i), \quad i = 1, 2, \dots, n.
\end{equation}
The rotated kernels are then used to convolve with the input feature map, and their outputs are combined in a weighted manner:

\begin{equation}
  Y = \sum_{i=1}^{n} \lambda_i (W^{\prime}_i * X).
\end{equation}

This approach improves the model’s ability to capture features from objects with varying orientations while maintaining computational efficiency. 

\section{Results and discussions}
\label{sec:5}

\subsection{Experimental Setting}

\subsubsection{Datasets}
\label{sec: datasets}

To evaluate the proposed method, we conducted extensive experiments on two newly developed UAV-RIS datasets, UAVid-RIS and VDD-RIS. Both datasets contain high-resolution images, all cropped to a size of 1024 × 1024 pixels.

\begin{table*}[]
\setlength{\abovecaptionskip}{0.cm}
\centering
    \caption{Performance comparison of different methods on UAVid-RIS. The table includes Precision at different IoU thresholds (Pr@0.5 to Pr@0.9), mean Intersection over Union (mIoU), overall Intersection over Union (oIoU), and the visual and textual encoders used in each method.}
    \label{tab:UAVid-ref}
    \begin{tabular}{l c c c c c c c c c}
        \hline
        Method  & Visual Encoder  & Textual Encoder  & Pr@0.5 & Pr@0.6 & Pr@0.7 & Pr@0.8 & Pr@0.9 & mIoU  & oIoU  \\
        \hline
        RIS-DMMI  & ResNet-101  & BERT    & 79.75  & 70.19  & 54.08  & 36.13  & 9.26   & 67.10 & 76.76 \\
        LAVT     & Swin-B      & BERT    & 84.45  & 75.75  & 59.60  & 38.71  & 10.91  & 69.32 & 78.76 \\
        LGCE    & Swin-B      & BERT    & 83.97  & 74.60  & 59.69  & 39.37  & 11.33  & 69.52 & 79.06 \\
        RMSIN    & Swin-B      & BERT    & 85.71  & 77.91  & 64.06  & 46.52  & 17.76  & 72.05 & 81.10 \\
        AeroReformer    & Swin-B      & BERT    & \textbf{86.34}  & \textbf{79.07}  & \textbf{65.60}  & \textbf{47.12}  & \textbf{18.52}  & \textbf{72.79} & \textbf{81.53} \\
        \hline
    \end{tabular}
\end{table*}

\begin{itemize}
  \item UAVid-RIS. This dataset consists of 7,035 images, divided into 3,215 for training, 1,163 for validation, and 2,657 for testing. UAVid \citep{lyu2020uavid} is designed for UAV-based scene understanding in complex urban environments, capturing both static and dynamic objects. The dataset features oblique-view aerial imagery with a camera angle of approximately 45 degrees, offering richer contextual information than nadir-view images. The data is collected from UAVs flying at an altitude of around 50 meters, with high-resolution frames extracted from 4K video recordings. The dataset includes diverse street scenes with objects such as vehicles, pedestrians, buildings, roads, vegetation, billboards, and traffic infrastructure.
  
  \item VDD-RIS. This dataset contains 1,941 images, split into 1,269 for training, 399 for validation, and 273 for testing. VDD \citep{cai2023vdd} is collected across 23 locations in Nanjing, China, covering diverse environments, including urban, rural, industrial, and natural landscapes. The dataset incorporates variations in camera angles, with images captured at 30, 60, and 90 degrees (nadir view), allowing for more comprehensive scene representation. The drone altitude ranges from 50 to 120 meters, ensuring a balance between scene complexity and fine-grained details. The dataset also introduces temporal and seasonal diversity, with images taken at different times of the day and in different seasons. 
\end{itemize}

\begin{table*}[]
\setlength{\abovecaptionskip}{0.cm}
\centering
    \caption{Performance comparison of different methods on VDD-RIS. The table includes Precision at different IoU thresholds (Pr@0.5 to Pr@0.9), mean Intersection over Union (mIoU), and overall Intersection over Union (oIoU).}
    \label{tab:VDD-ref}
    \begin{tabular}{l c c c c c c c c c}
        \hline
        Method  & Visual Encoder  & Textual Encoder  & Pr@0.5 & Pr@0.6 & Pr@0.7 & Pr@0.8 & Pr@0.9 & mIoU  & oIoU  \\
        \hline
        RIS-DMMI  & ResNet-101  & BERT    & 87.08  & 77.86  & 62.73  & 46.13  & 23.25   & 72.66 & 78.06 \\
        LAVT      & Swin-B      & BERT    & 90.04  & 86.72  & 76.01  & 56.09  & 34.32   & 77.80 & 82.51 \\
        LGCE     & Swin-B      & BERT    & 89.67  & 84.13  & 73.80  & 53.51  & 32.47   & 76.78 & 82.09 \\
        RMSIN    & Swin-B      & BERT    & 91.14  & 85.98  & 76.38  & 57.20  & 35.06   & 78.22 & 83.58 \\
        AeroReformer    & Swin-B      & BERT    & \textbf{92.99}  & \textbf{89.30}  & \textbf{81.55}  & \textbf{63.47}  & \textbf{38.75}  & \textbf{80.72} & \textbf{85.38} \\
        \hline
    \end{tabular}
\end{table*}

\subsubsection{Implementation Details}
\label{sec: details}

We implemented our method in PyTorch \citep{paszke2019pytorch}, utilising the pre-trained base BERT \citep{devlin2019bert} for language processing and the Swin Transformer \citep{liu2021swin} initialized with ImageNet-22K \citep{deng2009imagenet} weights for visual encoding.

All images were resized to 448 × 448 pixels, and no data augmentation (e.g., rotation, flipping) was applied. Training was conducted with a batch size of 8 for 40 epochs on UAVid-RIS and 10 epochs on VDD-RIS using the AdamW optimiser \citep{loshchilov2017decoupled} with a weight decay of 0.01 and an initial learning rate of 0.0005. Following the baseline LAVT \citep{yang2022lavt}, cross-entropy loss was used for optimisation. All experiments were performed on an NVIDIA RTX 5000 Ada GPU.

\begin{figure}[htb]
\centering
\includegraphics[width=8cm]{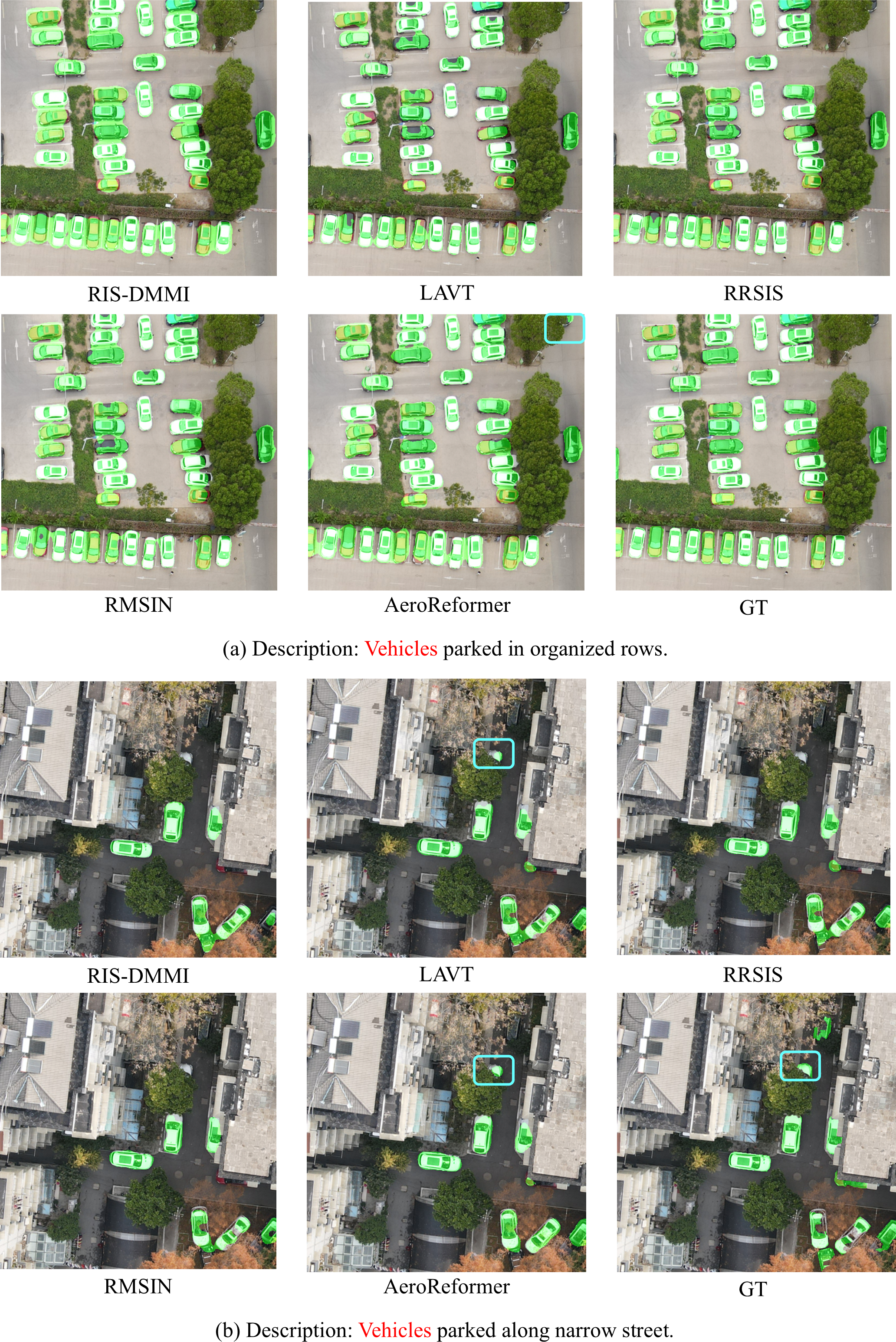}
\caption{Visual comparison of referring segmentation results on VDD-RIS dataset.}
\label{fig:8}
\end{figure}

\subsubsection{Model Evaluation}

For a fair comparison with previous methods \citep{yuan2024rrsis, liu2024rotated, yang2022lavt}, we adopted the same evaluation metrics, including mean Intersection over Union (mIoU), overall Intersection over Union (oIoU), and Precision at the 0.5, 0.6, 0.7, 0.8 and 0.9 IoU thresholds (Pr@X). 

The mIoU measures the average IoU between predicted and ground-truth masks across all test samples, giving equal weight to both large and small objects. In contrast, oIoU favours large objects by computing the ratio of the total intersection area to the total union area across all test samples. Additionally, Pr@X evaluates model performance at different IoU thresholds, reflecting the proportion of successfully predicted samples at each level. Higher values for these metrics indicate better segmentation performance.

\begin{table*}[]
\setlength{\abovecaptionskip}{0.cm}
\centering
    \caption{Class-wise Intersection over Union (IoU) and mean IoU (mIoU) for different methods on UAVid-RIS.}
    \label{tab:uavid-class}
    \begin{tabular}{l c c c c c c c c}
        \hline
        Method  & Building & Road & Tree & Low Vegetation & Moving Car & Static Car & Human & mIoU  \\
        \hline
        RIS-DMMI   & 84.48  & 77.91  & 79.67  & 64.08  & 67.04  & 50.18  & 24.11  & 63.92  \\
        LAVT       & 86.87  & 78.68  & 80.91  & 67.95  & 68.98  & 53.53  & 23.75  & 65.81  \\
        LGCE    & 87.58  & 79.36  & 80.67  & 67.97  & 67.78  & 58.07  & 24.01  & 66.49  \\
        RMSIN    & 89.02  & 84.33  & 82.01  & 69.23  & \textbf{72.31}  & 54.79  & 24.72  & 68.06  \\
        AeroReformer    & \textbf{89.12}  & \textbf{84.82}  & \textbf{82.35}  & \textbf{69.45}  & 72.14  & \textbf{60.58}  & \textbf{26.68}  & \textbf{69.31}  \\
        \hline
    \end{tabular}
\end{table*}

In addition to these metrics, we also report the class-wise IoU and class-wise mIoU to provide a more detailed analysis of segmentation performance across different object categories. Unlike object-based mIoU, which is calculated per test sample, class-wise mIoU is computed per class rather than per object, ensuring that the evaluation captures category-level segmentation accuracy instead of object-instance accuracy.

\subsection{UAV Referring Image Segmentation Performance}
In this section, we evaluate the performance of six different referring image segmentation (RIS) methods, including RIS-DMMI \citep{hu2023beyond}, LAVT \citep{yang2022lavt}, LGCE \citep{yuan2024rrsis}, RMSIN \citep{liu2024rotated}, and the proposed AeroReformer. The evaluation is conducted using standard segmentation metrics, including mean Intersection over Union (mIoU), overall Intersection over Union (oIoU), Precision at different IoU thresholds (Pr@X) and class-wise IoU. 

\subsubsection{Quantitative Results} 
\paragraph{Overall Performance} 
As shown in Tables \ref{tab:UAVid-ref} and \ref{tab:VDD-ref}, the proposed AeroReformer consistently achieves the highest scores across all evaluation metrics, demonstrating its superior segmentation capability. In UAVid-RIS, AeroReformer achieves an mIoU of 72.79 and an oIoU of 81.53, surpassing the second-best method, RMSIN, by 0.74 in mIoU and 0.43 in oIoU. Similarly, on VDD-RIS, AeroReformer achieves an mIoU of 80.72 and an oIoU of 85.38, outperforming RMSIN by 2.50 in mIoU and 1.80 in oIoU. These improvements highlight AeroReformer's effectiveness in capturing fine-grained segmentation details across different datasets.

\paragraph{Precision at Different IoU Thresholds}
In terms of precision at varying IoU thresholds, Tables \ref{tab:UAVid-ref} and \ref{tab:VDD-ref} illustrate that AeroReformer consistently outperforms the second-best method across all threshold levels. On UAVid-RIS, AeroReformer achieves the highest Pr@0.5 score of 86.34, surpassing RMSIN by 0.63, and maintains its lead at Pr@0.9 with 18.52, exceeding RMSIN by 0.76. On VDD-RIS, AeroReformer achieves a Pr@0.5 of 92.99, improving upon RMSIN by 1.85, and maintains the best performance at Pr@0.9 with 38.75, surpassing RMSIN by 3.59. These improvements confirm AeroReformer's robustness and reliability in maintaining segmentation accuracy under different IoU thresholds.

\begin{table*}[]
\setlength{\abovecaptionskip}{0.cm}
\centering
    \caption{Class-wise Intersection over Union (IoU) and mean IoU (mIoU) for different methods on VDD-RIS.}
    \label{tab:vdd-classv}
    \begin{tabular}{l c c c c c c c}
        \hline
        Method  & Wall & Road & Vegetation & Vehicles & Roof & Water & mIoU  \\
        \hline
        RIS-DMMI   & 65.95  & 77.21  & 89.43  & 66.38  & 76.77  & 80.63  & 76.06  \\
        LAVT       & 72.62  & 82.94  & 90.04  & 69.07  & 81.01  & \textbf{91.30}  & 81.16  \\
        LGCE     & 70.63  & 80.75  & 89.97  & 69.42  & 82.09  & 90.03  & 80.48  \\
        RMSIN    & 73.16  & \textbf{82.64}  & 89.39  & 70.11  & 84.88  & 90.54  & 81.79  \\
        AeroReformer    & \textbf{76.82}  & 82.57  & \textbf{91.78}  & \textbf{70.74}  & \textbf{86.21}  & 91.25  & \textbf{83.23}  \\
        \hline
    \end{tabular}
\end{table*}

\paragraph{Class-wise IoU Analysis}  
A deeper analysis of class-wise IoU scores in Tables \ref{tab:uavid-class} and \ref{tab:vdd-classv} further supports AeroReformer's superior performance. On UAVid-RIS, AeroReformer achieves the highest IoU scores in six out of seven categories. Compared to the second-best method, RMSIN, AeroReformer improves static car segmentation by 5.79 and human segmentation by 1.96. RMSIN slightly outperforms AeroReformer in the moving car category, but AeroReformer still maintains the highest overall mIoU. On VDD-RIS, AeroReformer achieves the highest IoU scores in five out of six categories. It surpasses second best by 3.66 in wall segmentation and 0.63 in vehicle segmentation. LAVT outperforms AeroReformer in the water category by 0.05. Despite this, AeroReformer achieves the highest overall mIoU of 83.23, improving by 1.44.

\subsubsection{Qualitative Results}

To further evaluate the segmentation performance of different RIS methods, we present qualitative comparisons on UAVid-RIS and VDD-RIS in Fig.~\ref{fig:7} and \ref{fig:8}. Each example consists of segmentation results from five different methods: RIS-DMMI, LAVT, RRSIS, RMSIN, and the proposed AeroReformer, along with the ground truth (GT). The visualized results highlight AeroReformer's ability to produce more precise and contextually accurate segmentations.

\paragraph{Results on UAVid-RIS}  
Fig. \ref{fig:7} illustrates segmentation results for two different referring expressions: \textit{``Vehicles parked near houses"} and \textit{``Vehicles move along the road"}. In the first example, all methods correctly segment the parked vehicles in the centre of the image. However, RIS-DMMI, LAVT, and RRSIS mistake part of the cottage as a vehicle, leading to incorrect segmentation. AeroReformer provides the most consistent segmentation, aligning closely with the ground truth. In the second example, which describes moving vehicles, RIS-DMMI struggle to differentiate parked cars with moving vehicles, while RMSIN misidentifies electric bikes as motor vehicles. AeroReformer provides the most precise segmentation, accurately detecting moving cars along the road while avoiding the misclassification of static objects. These results demonstrate AeroReformer's improved ability to distinguish between stationary and dynamic objects in UAV imagery.

\paragraph{Results on VDD-RIS}  
Fig.~\ref{fig:8} presents segmentation results for two different referring expressions: \textit{``Vehicles parked in organized rows"} and \textit{``Vehicles parked along a narrow street"}. In the first example, all methods perform well in detecting parked vehicles. Notably, in the top right corner, a black car is partially covered under the trees; even though the ground truth does not label it, AeroReformer successfully detects it, demonstrating its superior ability to capture occluded objects. In the second example, which depicts vehicles parked along a narrow street, RIS-DMMI, RRSIS, and RMSIN fail to distinguish a parked car from surrounding trees, particularly when a car is partially covered by foliage. Only LAVT and AeroReformer accurately differentiate the vehicles, minimising misclassification. This demonstrates AeroReformer's strong performance in complex urban environments with occlusions and varying object scales.

\paragraph{Summary}  
The qualitative comparisons across UAVid-RIS and VDD-RIS confirm that AeroReformer produces more precise and contextually accurate segmentations than existing methods. It consistently outperforms the second-best method by correctly distinguishing between similar objects and capturing finer details. These results validate AeroReformer's effectiveness in complex aerial scenes with dynamic and static objects.

\subsection{Ablation study} 

\begin{table}[]
\centering
\caption{Ablation study results on the VDD-RIS dataset. The table evaluates the impact of RAMSF and VLCAM, showing Precision at IoU thresholds Pr@0.5, Pr@0.7, and Pr@0.9, and mean Intersection over Union (mIoU). A checkmark (\checkmark) indicates the inclusion of a module.}
\label{tab:ablation}
\begin{tabular}{c c c c c c}
\hline
RAMSF & VLCAM & Pr@0.5 & Pr@0.7 & Pr@0.9 & mIoU  \\
\hline
--    & \checkmark  & 90.04  & 78.60  & 36.53  & 78.27  \\
\checkmark  & --    & 91.14  & 77.86  & 36.16  & 78.83  \\
\checkmark  & \checkmark  & 93.01  & 83.46  & 37.13  & 80.88  \\
\hline
\end{tabular}
\end{table}

To analyse the contribution of each proposed module in AeroReformer, we conduct an ablation study on the VDD-RIS dataset. Table~\ref{tab:ablation} presents the results when removing the Rotation-Aware Multi-Scale Fusion (RAMSF) decoder and the Vision-Language Cross Attention Module (VLCAM). The evaluation is performed Pr@0.5, Pr@0.7, Pr@0.9 and mIoU.

\paragraph{Effect of RAMSF}  
The first row in Table~\ref{tab:ablation} shows the performance when RAMSF is removed while keeping VLCAM. Compared to the full model, removing RAMSF leads to a performance drop of 2.97 in Pr@0.5, 4.86 in Pr@0.7, and 0.60 in Pr@0.9. Additionally, mIoU decreases by 2.61. These results confirm that RAMSF plays a crucial role in enhancing spatial consistency and preserving fine details across different scales.

\paragraph{Effect of VLCAM}  
The second row in Table~\ref{tab:ablation} reports the results when VLCAM is removed while retaining RAMSF. The performance drop is 1.87 in Pr@0.5, 5.60 in Pr@0.7, and 0.97 in Pr@0.9, with a decrease of 2.05 in mIoU. This indicates that VLCAM is essential for refining cross-modal interactions and improving segmentation precision, particularly at higher IoU thresholds.
 
\paragraph{AeroReformer} The complete AeroReformer, which integrates both RAMSF and VLCAM, achieves the highest scores across all metrics, with an mIoU of 80.88 and significant improvements at all IoU thresholds. The ablation study confirms that both RAMSF and VLCAM contribute significantly to AeroReformer's performance. RAMSF enhances spatial feature fusion, while VLCAM strengthens visual-linguistic alignment. Their combination leads to consistent improvements across all metrics, proving their necessity for high-quality referring image segmentation in aerial imagery.

\section{Conclusions}
\label{sec:6}

In this work, we proposed a fully automated dataset construction pipeline that transforms pre-existing UAV segmentation datasets into referring segmentation benchmarks. The designed pipeline leverages segmentation masks and large language models to generate diverse and contextually accurate referring expressions. This method was applied to UAVid and VDD, producing UAVid-RIS and VDD-RIS, two novel datasets that expand the applicability of vision-language segmentation in UAV imagery. In addition, we introduced AeroReformer, a novel framework for referring image segmentation that integrates the Rotation-Aware Multi-Scale Fusion (RAMSF) decoder and the Vision-Language Cross-Attention Module (VLCAM) to enhance spatial feature fusion and cross-modal alignment. AeroReformer consistently outperforms comparative methods on UAVid-RIS and VDD-RIS, demonstrating superior segmentation accuracy in challenging UAV environments with occlusions, scale variations, and fine-grained object details. The ablation study further validates the necessity of RAMSF and VLCAM, showing that their combination significantly boosts segmentation performance.

While AeroReformer achieves significant improvements, it still relies on a separate vision encoder for segmentation. Future work should explore the integration of segmentation capabilities directly into large language models (LLMs), eliminating the need for external vision modules. 

\section*{Declaration of Competing Interest}
\par The authors declare that they have no known competing financial interests or personal relationships that could have appeared to influence the work reported in this paper.

\section*{CRediT authorship contribution statement}
\textbf{Rui Li}: Formal analysis, Conceptualization, Investigation, Methodology, Project administration, Software, Validation, Visualization, Writing - original draft. \textbf{Xiaowei Zhao}: Project administration, Resources, Supervision, Writing - review and editing.

\section*{Acknowledgements}
This work has received funding from the UK Engineering and Physical Sciences Research Council (grant number: EP/Y016297/1).

\bibliographystyle{apalike}
\bibliography{AeroReformerAbbv}
\end{document}